\documentclass{article}

\PassOptionsToPackage{numbers, compress}{natbib}

 \usepackage[main, final]{neurips_2026}

\usepackage[utf8]{inputenc} 
\usepackage[T1]{fontenc}    
\usepackage{hyperref}       
\usepackage{url}            
\usepackage{booktabs}       
\usepackage{amsfonts}       
\usepackage{nicefrac}       
\usepackage{microtype}      
\usepackage{xcolor}         

\usepackage{epsfig}
\usepackage{graphicx}
\usepackage{subcaption}
\usepackage{caption}
\usepackage{tabularx}
\usepackage{makecell}
\usepackage{wrapfig}
\usepackage{amsmath}
\usepackage{amssymb}
\usepackage{multicol}
\usepackage{multirow}
\usepackage{bbding}
\usepackage{dsfont}

\usepackage{algorithm}
\usepackage{algorithmic}
\title{Marrying Optimal Transport and ODEs for Unified Continuous-Time 4D Reconstruction and Tracking}

\author{%
  Liying Yang$^{1}$, Hao Mo$^{1}$, Jialun Liu$^{2\dagger}$, Chen Liu$^{2}$, Xinxing Yu$^{1}$, Chenhao Guan$^{1}$, \\
  \textbf{Hui Ma$^{3,4}$, Xiao Cao$^{5}$, Ajian Liu$^{6}$, Yanyan Liang$^{1\dagger}$} \\
  School of Computer Science and Engineering, Macau University of Science and Technology$^{1}$ \\ The University of Queensland$^{2}$ \qquad
  Great Bay University$^{3}$ \qquad Tsinghua University$^{4}$ \\ National University of Singapore$^{5}$ \qquad
  Institute of Automation, Chinese Academy of Sciences$^{6}$\\
  Corresponding Authors$^{\dagger}$ \\
}

\begin{document}
\maketitle

\begin{abstract}
Existing unified 4D reconstruction and point tracking approaches typically rely on heuristic interpolations or just predict at integer timestamps, lacking kinematic coherence and failing to model dynamics at any arbitrary timestamp. In this paper, we propose Uni4R, a framework that unifies these tasks by learning continuous velocity fields through the synergy of Optimal Transport (OT) and Ordinary Differential Equation (ODE). Importantly, this continuous velocity field acts as a kinematic prior that mutually benefits both 4D reconstruction and point tracking. Specifically, we propose the Flow Matching Guided Decoder (FMGD). A global velocity branch first extracts anchor features that capture the global dynamic state of the sequence. Then, FMGD leverages Flow Matching (FM) theory to formulate a probability path defined by OT on the anchor feature manifold, instantiating it as FM-guided velocity features for velocity prediction. This establishes a robust kinematic inductive bias. Meanwhile, a point reconstruction branch provides geometric features. The local velocity prediction module then joint above features and time embeddings, to decode velocities at arbitrary timestamps. To overcome the absence of high-quality ground-truth velocities in fractional frames, we propose an integral-consistency training strategy. This strategy uses an ODE solver to integrate velocities to recover target pointmaps, enabling the model to be supervised end-to-end directly from integer timestamps. Experimental results demonstrate that Uni4R achieves SOTA performance in both 4D reconstruction and point tracking, and achieves SOTA in our new kinematics-aware benchmark at continuous time. Project page: \texttt{https://LiyingCV.github.io/Uni4R/\href{https://LiyingCV.github.io/Uni4R/}}
\end{abstract}

\section{Introduction}
Existing unified 4D reconstruction and point tracking methods~\cite{feng2025st4rtrack,sucar2026v} mainly focus on integer time steps, which restricts their temporal resolution. To establish point correspondences at arbitrary timestamps, some recent methods~\cite{liu2025trace} resort to heuristic interpolations (e.g., B-spline). Unfortunately, these approximations lack underlying kinematic coherence, which requires points to propagate along continuous and energy-minimizing trajectories. By treating motion modeling as mere curve fitting rather than a continuous geometric flow, they frequently fail to preserve local geometric structures, often leading to distortions. More importantly, without a continuous kinematic prior, these methods fail to fully exploit the inherent mutual benefits between reconstruction and tracking.

Formulating 4D dynamics as an Ordinary Differential Equation (ODE)~\cite{chen2018neural,dupont2019augmented,norcliffe2020second} local initial value problem offers a principled paradigm to achieve true spatiotemporal continuity. Given that ODE naturally yield smooth trajectories, we leverage ODE to learn the continuous velocity field. Furthermore, this continuous velocity field acts as a kinematic priors that seamlessly bridges spatial topology and temporal dynamics. \textit{Under this prior constraint, accurate point tracking provides kinematic guidance for geometric deformation, while high-quality 4D reconstruction regularizes the trajectory space, allowing the two tasks to mutually enhance each other's quality.} However, realizing this continuous formulation poses an optimization dilemma that when supervising a continuous velocity field using only point maps at integer timestamps, the inverse problem becomes highly ill-posed. Infinitely many arbitrary, curved velocity fields transport a point from state $\mathbf{P}_t$ to $\mathbf{P}_{t+1}$, rendering the model prone to overfitting and resulting in unstable trajectories.

\begin{figure}[t]%
\centering
\includegraphics[width=1.0\textwidth]{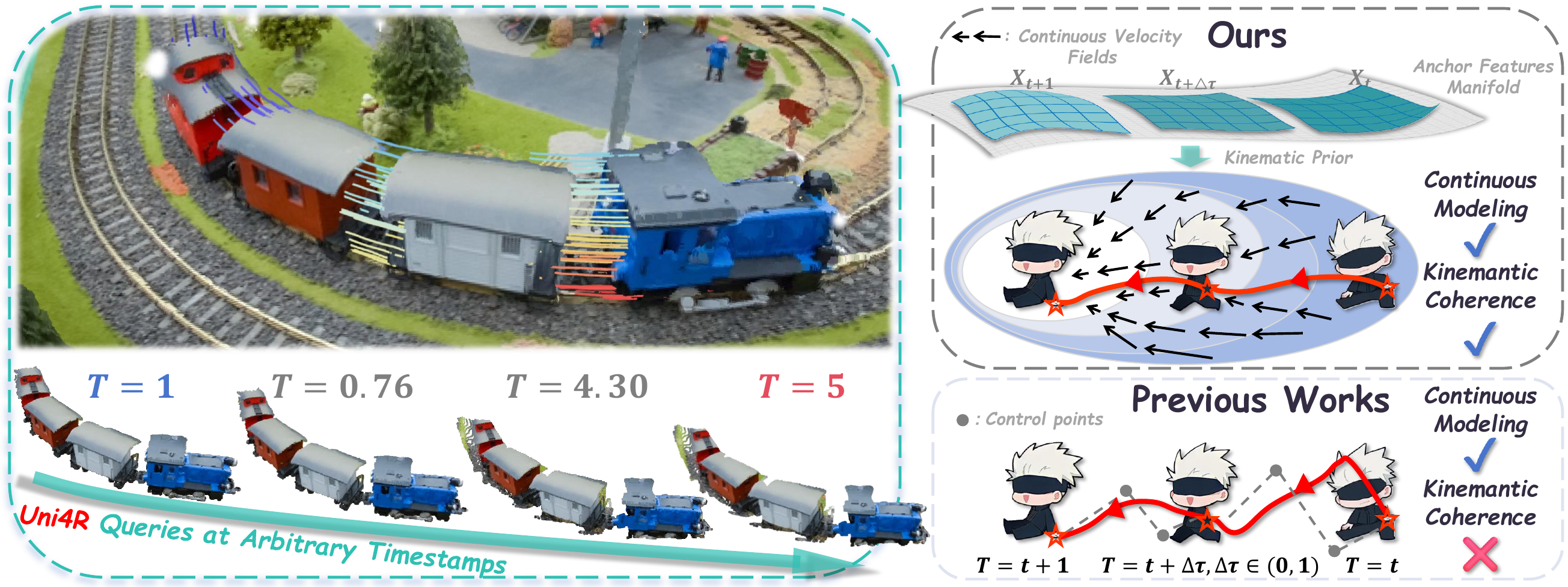}
\caption{\textbf{Left:} Uni4R is a feed-forward transformer that queries point maps at arbitrary timestamp. \textbf{Right:} While existing methods like TraceAnything~\cite{liu2025trace} rely on B-spline fitting and inherently lack kinematic coherence,  Uni4R learns continuous velocity fields via the synergy of OT and ODE. Specifically, we formulate a probability path defined by OT on the anchor feature manifold to provide a kinematic inductive bias. Then, an ODE solver integrates velocities queried from the continuous velocity field to achieve continuous modeling with kinematic coherence.}\label{teaser}
\end{figure}


To overcome these limitations, we propose Uni4R, a unified framework establishing a deep synergy between Optimal Transport (OT) and Ordinary Differential Equations (ODEs). Within this synergy, ODEs provide the mathematical foundation for continuous motion prediction at any arbitrary time, while OT provides a kinematic inductive bias with a global velocity direction prior to constrain the ill-posed problem. At its core, our proposed Flow Matching Guided Decoder (FMGD) leverages Flow Matching (FM) theory~\cite{lipman2022flow} to formulate the probability path defined by OT on anchor feature manifold. Specifically, a global velocity branch first extracts anchor features that capture the global dynamic state of the input sequence, implicitly embedding motion information for any non-integer moment. By formalizing this OT-defined probability path within the feature manifold, we obtain FM-guided velocity features that encode a global velocity direction prior as a kinematic inductive bias. Furthermore, a point reconstruction branch extracts geometric features. These geometric features and the FM-guided velocity features, conditioned on continuous time embeddings, are then sent to our local velocity prediction module to decode velocities at arbitrary timestamps. Through this unified framework, FMGD effectively constrains the ill-posed continuous inverse problem, enabling smooth and kinematically consistent 4D trajectories.

While FMGD can decode velocities, training this velocity field is hindered by the inherent absence of high-quality ground-truth velocities in fractional frames. To bridge the gap between predicted dynamics and integer-time ground-truth, we propose an integral-consistency training strategy. By employing an ODE solver to integrate velocities to recover target pointmaps, we enable robust end-to-end supervision from integer pointmaps. Ultimately, within this unified framework, the continuous velocity field acts as a kinematic prior that mutually benefits the accuracy of two tasks by implicitly modeling continuous motion. Remarkably, despite employing ODE, given 80-frame inputs, Uni4R requires only 1.57 seconds to infer results (66.52\% faster than Trace Anything~\cite{liu2025trace}).

Current evaluation protocols for point tracking mainly focus on integer timestamp, lacking standard evaluation in continuous time. To bridge this gap, we introduce a kinematics-aware benchmark designed to systematically evaluate kinematic coherence at continuous time.

In summary, the main contributions of our work are as follows:
\begin{enumerate} 
\vspace{-1pt}
\item We propose Uni4R, a unified framework that shifts 4D reconstruction and point tracking from heuristic interpolations to continuous kinematic modeling, unlocking the capability to query dynamics at any arbitrary timestamp.
\vspace{-1pt}
\item We propose Flow Matching Guided Decoder (FMGD), which leverages OT to provide a global velocity direction prior as a kinematic inductive bias for constraining the ill-posed inverse problem. Moreover, we design an integral-consistency training strategy, enabling end-to-end supervision of continuous velocity field from integer time steps through ODE.

\vspace{-1pt}
\item We present a kinematics-aware benchmark that establishes a unified evaluation standard for kinematic coherence in continuous time.
\vspace{-1pt}
\item Experimental results demonstrate that Uni4R achieves SOTA performance in both 4D reconstruction and point tracking, and achieves SOTA in kinematics-aware benchmark.
\end{enumerate}

\section{Related Works}

\textbf{Reconstruction and Generation.} Creating high-quality 3D and 4D contents, such as voxels~\cite{xie2019pix2vox,zhu2023garnet,yang2023long,zhu2023umiformer},meshes~\cite{long2024wonder3d,zhao2025hunyuan3d,hunyuan3d2025hunyuan3d}, point clouds~\cite{melas2023pc,yu2026pointmc,yu2025facnet,yu2026pointcsp,yu2026pointchr}, and 3D Gaussian splats~\cite{kerbl20233d,zhang2024gs,tang2024lgm} has become a rapidly evolving research area. Recent 4D generation methods~\cite{zheng2024unified,zhao2023animate124,zeng2024stag4d,wu2024sc4d,yang2025not} employ dynamic or multi-view prior from video diffusion~\cite{singer2022make,wan2025wan} or multi-view diffusion~\cite{liu2023zero,shi2023zero123++} to optimize dynamic 4D representations via Score Distillation Sampling (SDS)~\cite{poole2022dreamfusion,lyu2026choreographing}. Recent methods~\cite{ren2024l4gm,yuan2025seeu,yao2025sv4d,yang2026spatial} generate 4D gaussian splatting in feed-forward manner based on diffusion or autoregressive model. Different from them, recent 3D and 4D reconstruction~\cite{wang2024dust3r,yang2025fast3r,wang2025vggt,wang2025pi,sucar2025dynamic,liu2025trace,sucar2026v} output disjoint per-frame pointmaps instead of gaussian splatting~\cite{kerbl20233d,wang2026mcgs}. However, most of them focus on integer time steps. In contrast, Uni4R commits to modeling dynamics both at integer and fractional time steps.

\textbf{Point Tracking.} The paradigm of motion estimation has rapidly evolved from 2D optical flow to 3D point tracking. To enable tracking through occlusions, CoTracker~\cite{karaev2024cotracker} employs a transformer-based architecture, while CoTracker3~\cite{karaev2025cotracker3} leveraged unlabeled data to boost performance. SpatialTracker \cite{xiao2024spatialtracker} proposes the first feed-forward 3D tracker though  combining 2D tracking with monocular depth priors. SpatialTrackerV2 \cite{xiao2025spatialtrackerv2} scaled training across real and synthetic data. Recently, 3D tracking-based methods~\cite{feng2025st4rtrack,zhang2025pomato,liu2025trace,sucar2026v,qian2026flow4r,zhang2025efficiently} has shown impressive results. St4RTrack \cite{feng2025st4rtrack} and POMATO \cite{zhang2025pomato} demonstrate the benefits of jointly predict 4D reconstruction and tracking. Track4World \cite{lu2026track4world} applies a 3D correlation scheme to estimate dense 2D and 3D tracking. TraceAnything \cite{liu2025trace} predicts a set of control points that parameterizes a trajectory via B-spline. However, they frequently fail to preserve local geometric structures due to lack underlying kinematic coherence. In contrast, Uni4R models dynamics via kinematic inductive both at integer and fractional time steps.

\textbf{Flow Matching.} Continuous Normalizing Flows (CNFs)~\cite{chen2018continuous,grathwohl2018ffjord} have emerged as powerful tools for modeling distributions via Ordinary Differential Equations (ODEs). The pioneering work \cite{lipman2022flow} established a simulation-free training paradigm for continuous vector fields and was improved by \cite{tong2023improving}. \cite{tong2023improving} integrates Optimal Transport (OT) to constrain ODE trajectories into deterministic, minimum-energy linear paths. However, its potential as a kinematic inductive bias for solving point tracking and 4D reconstruction remains unexplored. Our Uni4R formulates probability path defined by OT not for generation, but to query continuous and kinematics-aware velocities across feature manifold.

\section{Methodology}
\begin{figure}[t]%
\centering
\includegraphics[width=1.0\textwidth]{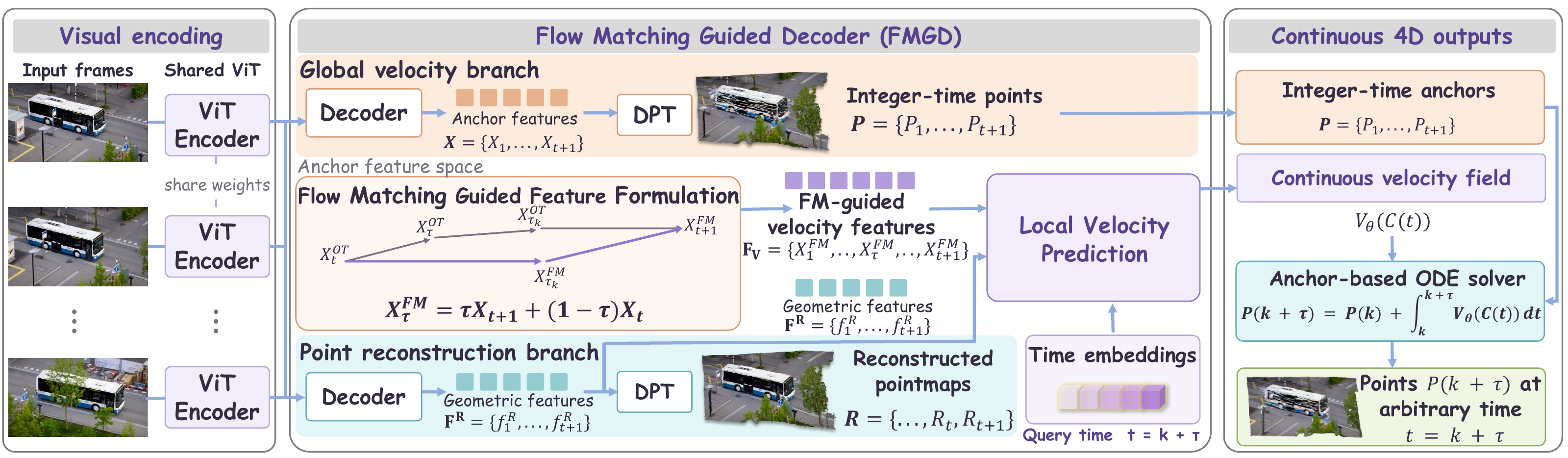}
\caption{\textbf{Pipeline of Uni4R.} A ViT encoder extracts visual features from input frames, which are processed into geometric features $F^R$ and anchor features $X$. Then, we formulate FM-guided velocity features $F_V$ via an OT probability path on the anchor feature manifold. This provides a kinematic inductive bias characterizing feature-space transport. Finally, a Local Velocity Prediction (LVP) module fuses $X, F^R, F_V$, and time embeddings to predict a continuous velocity field, which an ODE solver integrates to produce points at any arbitrary time $t$.}\label{pipeline}
\end{figure}

Given a monocular frame sequence of a dynamic scene, Uni4R first extracts visual features using a shared ViT encoder. These features are then processed by FMGD for unified 4D reconstruction and point tracking. Fig.~\ref{pipeline} shows the overview of Uni4R.

\subsection{Continuous 4D Dynamics Modeling via Ordinary Differential Equation} \label{method_ode}

Formally, to achieve continuous dynamic modeling, we model the 4D dynamics as a sequence of local initial value problems. Instead of integrating from a single global origin (i.e., global-based ODE), we define $P(k)$ as the anchor position at integer timestamp $k$. The position of a point at any arbitrary timestamp $k+\tau$ (where $\tau \in [0, 1)$) is then determined by integrating the velocities queried from continuous velocity field $\mathbf{V}_{\theta}(C(t))$ starting from its nearest anchor $P(k)$, named anchor-based ODE, where $C(t)$ denotes the time-dependent conditioning information of the velocity predictor:

\begin{equation}
P(k+\tau) = P(k) + \int_{k}^{k+\tau} \mathbf{V}_{\theta}(C(t)) dt
\end{equation}

By grounding the continuous velocity field on these periodic anchors, our framework effectively mitigates the accumulation of integration errors. However, recovering the continuous velocity field solely from observations at discrete integer timestamps remains a highly ill-posed inverse problem. While the ODE formulation ensures that a given velocity field yields a deterministic trajectory, there exists an infinite number of continuous paths that can align with the ground truth at integer timestamps. Therefore, our framework requires a robust kinematic inductive bias to constrain the solution space and predict the most kinematically consistent trajectory among all possible continuous paths.

\begin{wrapfigure}{r}{0.3\textwidth} %

  \centering
  \vspace{-15pt} %
  
  \includegraphics[width=0.75\linewidth]{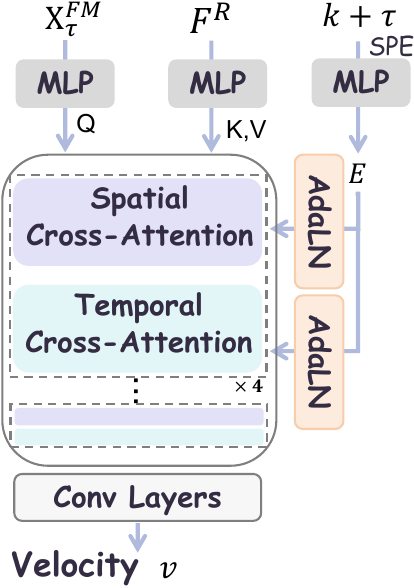} %
  
  \caption{The details of LVP.}
  \label{fig:lvp}
  
  \vspace{-18pt} %
\end{wrapfigure}

\subsection{Flow Matching Guided Decoder} \label{method_fmgd}
To mitigate above ill-posed problem, we propose the Flow Matching Guided Decoder (FMGD). FMGD leverages a kinematic inductive bias based on Optimal Transport (OT). This design establishes the synergy of OT and ODE. \textbf{The ODE enables continuous motion prediction at arbitrary times, while OT provides a global velocity direction prior to constrain the ill-posed problem.}

\textbf{Flow Matching Guided Feature Formulation:} According to the Flow Matching theorem for OT, the efficient probability path between a source state $x_0$ and a target state $x_1$ is a straight line~\cite{lipman2022flow}. For any arbitrary time $t$, we define the target conditional latent velocity field $u_t$ to construct this optimal path as $u_t(x|x_0, x_1) = x_1 - x_0$. This formulation minimizes kinetic energy~\cite{benamou2000computational} and significantly reduces numerical truncation errors during continuous integration~\cite{liu2022flow}. The continuous latent velocity field $v_\theta$ is thus optimized via the conditional flow matching loss:

\begin{equation}
\mathcal{L}_{CFM}(\theta) = \mathbb{E}_{t, q(x_0, x_1), p_t(x|x_0, x_1)} \left[ \| v_\theta(x, t) - u_t(x|x_0, x_1) \|_2^2 \right]. \label{flow_func}
\end{equation}

To learn continuous latent velocity field $v_\theta$ within FMGD, we propose a local velocity prediction (LVP) module (Fig.~\ref{fig:lvp}). Inspired by~\cite{feng2025st4rtrack}, we decouple coarse motion and geometry using two independent decoder-DPT pairs ~\cite{wang2024dust3r}, including a global velocity branch and a point reconstruction branch. 

From global velocity branch, we extract anchor features $\mathbf{X} \in \mathbb{R}^{B \times T \times N \times D}$ at integer time $1$ to $T+1$. These anchor features capture the global dynamic state of the sequence, implicitly embedding motion information for any non-integer moment. Guided by the optimal transport (OT) prior for minimum-energy paths ~\cite{tong2023improving,benamou2000computational}, we construct FM-guided velocity features at any fractional timestamp $\tau \in (0,1)$ via convex combination: $\mathbf{X}_{\tau}^{\text{FM}} = \tau\mathbf{X}_1 + (1-\tau)\mathbf{X}_2$. We randomly sample a set $\{\tau\}$ in training, while $\{\tau\}$ set is determined based on the specified integration step size in inference. Specifically, $\mathbf{X}_2=\{X_1,\dots,X_{T}\}$ and $\mathbf{X}_1=\{X_2,\dots,X_{T+1}\}$ act as source state $x_0$ and target state $x_1$ defined in \eqref{flow_func}. This defines a probability path that minimizes the OT cost~\cite{lipman2022flow} to yield kinematically consistent features. Note that this OT-defined path lies in the anchor feature space and does not directly represent the 3D point trajectory. The point reconstruction branch extracts geometric features $\mathbf{F^R} \in \mathbb{R}^{B \times T \times N \times D}$ (decoded into pointmaps via DPT) to provide geometric context for velocity prediction.

To decode velocities for an arbitrary fractional sequence of length $L$ (equal to the size of the set $\{\tau\}$), the arbitrary time values $k+\tau$ ($k$ is any integer time) are embedded into $\mathbf{E} \in \mathbb{R}^{B \times L \times d}$ via sinusoidal positional encodings (SPE) and an MLP. We then utilize $\mathbf{X}_{\tau}^{\text{FM}}$ as queries to attend to the geometric context $\mathbf{F^R}$ via a Spatial-Temporal Cross-Attention module, accelerated by FlashAttention~\cite{dao2022flashattention}. We perform spatial cross-attention along spatial axes while temporal cross-attention along time axis. Prior to each attention layer, queries are conditioned on $\mathbf{E}$ via Adaptive Layer Normalization (AdaLN) to inject continuous temporal awareness. Here, spatial attention corrects local trajectories against the global geometry, while temporal attention enforces smoothness across all timestamps.

Finally, the LVP outputs velocity field features $\mathbf{F_\text{vel}} \in \mathbb{R}^{B \times L \times T \times N \times D}$. We perform the CFM loss between $\mathbf{X}_{1}-\mathbf{X}_{2}$ (as the value of $u_t(\cdot)$) and $\mathbf{F_\text{vel}}$ (as the value of $v_\theta(\cdot)$) via \eqref{flow_func}. Moreover, convolution layers decode $\mathbf{F_\text{vel}}$ into continuous 3D velocity field $\mathbf{V}_{\theta}(C(t))$, where $C(t)$ consists of the corresponding FM-guided velocity features, geometric features, and time embeddings. This allows us to query 3D velocities at integer and fractional timesteps. By conditioning FM-guided velocity features on geometric features and $\mathcal{L}_{CFM}$, the learned velocity field adheres to the underlying geometric structure. Integrating this structurally constrained field yields smooth, kinematically coherent trajectories. The reconstruction regularizes this field, while the resulting point tracking provides reliable motion guidance for reconstruction. This unified framework ultimately drives two tasks to mutually enhance each other's quality.

\subsection{Integral-Consistency Training Strategy} \label{ic_loss}

Obtaining high-quality ground truth velocity for fractional timestamps is inherently difficult. While discrete anchors mitigate long-range drift, integrating velocities without direct supervision leaves the trajectory vulnerable to positional discontinuity (i.e., lacking $C^0$ continuity~\cite{farin2002curves}) at integer boundaries. We address this by proposing an integral-consistency training strategy. Given integer timestep $T$, we take the pointmaps predicted by DPT at $T$, and integrate them forward to $T+1$ using velocity field $\mathbf{V}_{\theta}(C(t))$. Specifically, $P(t)$ denotes the point trajectory state obtained during ODE integration. We solve the ODE using the efficient first order Euler method, which is highly effective because our kinematic inductive bias based on OT naturally reduces discretization errors~\cite{liu2022flow,lipman2022flow}. Next, the integrated pointmaps is then supervised by the GT pointmaps at $T+1$. The process is formulated as follows:

\begin{equation}
\mathcal{L}_{ICT} = \mathbb{E}_{T, P} \left[ \left\| P_{T+1} - \left( P_T + \int_{T}^{T+1} \mathbf{V}_{\theta}(C(t)) dt \right) \right\|_2^2 \right]
\end{equation}

In the absence of intermediate velocity supervision, this training strategy bridges the gap by translating unobservable velocity fields into explicitly supervised pointmaps. Importantly, the loss also backpropagates through integration path directly into the upstream DPT and Decoder. This gradient flow explicitly regularizes the accuracy of the predicted pointmaps and velocities at integer timestamp.

\textbf{Total Loss Function:} Following DUSt3R~\cite{wang2024dust3r}, we also employ DUSt3R's loss functions. The total loss is $\mathcal{L}=\lambda_1\mathcal{L}_{ICT}+\lambda_2\mathcal{L}_{CFM}+\lambda_3\mathcal{L}_{DUSt3R}$, where $\lambda_1:\lambda_2:\lambda_3=1:1:1$.

\begin{table}[t]
  \centering

  \renewcommand{\arraystretch}{1.13}
  \setlength{\tabcolsep}{6pt}
  
  \captionof{table}{\textbf{World Coordinate 3D Point Tracking.} We report the performance on datasets PO, DR, ADT, and PStudio using the APD ($\text{APD}_\text{3D}\uparrow$) after global median alignment. We evaluate the accuracy of both all points and dynamic points. All the experiments of methods are carried out using the code and pre-trained model from their official GitHub. The best results are \textbf{bold}.}
  \label{tab:3dtracking}
  
  \resizebox{1\textwidth}{!}{
    \begin{tabular}{@{}llcccccccc@{}}
    \toprule
     & & \multicolumn{4}{c}{All Points} & \multicolumn{4}{c}{Dynamic Points} \\
    \cmidrule(lr){3-6} \cmidrule(lr){7-10}
    \textbf{Category} & \textbf{Methods} & PO & DR & ADT & PStudio & PO & DR & ADT & PStudio \\
    \midrule
    \multirow{2}{*}{\textbf{Combinational}} 
      & SpaTracker+RANSAC-Procrustes 
        & 44.03 & 55.01 & 50.87 & 52.05 & 53.77 & 58.58 & 66.49 & 52.05 \\
      & SpaTracker+MonST3R
        & 47.65 & 55.49 & 51.95 & 50.16 & 58.61 & 59.21 & 69.94 & 50.16 \\
    \midrule
    \multirow{5}{*}{\textbf{Feed-forward}}
      & MonST3R 
        & 33.47 & 58.06 & 74.35 & 51.32 & 39.36 & 51.86 & 67.92 & 51.32 \\
      & SpaTracker 
        & 38.54 & 54.85 & 45.65 & 62.59 & 51.20 & 58.65 & 67.65 & 62.59 \\
      & POMATO
        & 49.73 & 68.36 & 57.10 & 64.91 & 58.14 & 62.84 & 78.12 & 64.87 \\
      & St4RTrack 
        & 67.95 & 73.74 & 76.01 & 69.67 & 68.72 & 68.13 & 75.34 & 69.67 \\
      & TraceAnything
        & 67.58 & 73.09 & 74.72 & 70.34 & 71.86 & 75.96 & 76.14 & 70.34 \\
      & V-DPM
        & 81.27 & 75.32 & 85.72 & 78.76 & 82.57 & 78.28 & 73.45 & 78.76 \\
        \cmidrule{2-10}
      & \textbf{Uni4R (Ours)} 
        & \textbf{81.59} & \textbf{76.05} & \textbf{86.55} & \textbf{79.43}
        & \textbf{83.16} & \textbf{78.91} & \textbf{77.74} & \textbf{79.43} \\
    \bottomrule
    \end{tabular}
  }

  \vspace{-1pt} %

  \begin{minipage}[t]{0.47\textwidth}
  \vspace{1.5em}
    \centering
    \includegraphics[width=\linewidth]{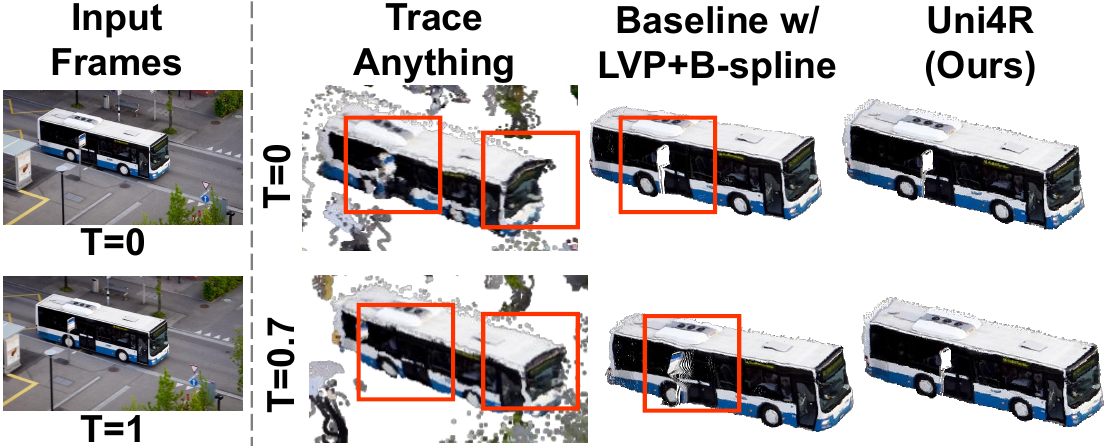} 
    \vspace{2pt}
    \captionof{figure}{\textbf{Comparison with Continuous Modeling Variant and Trace Anything}~\cite{liu2025trace} \textbf{at fractional timestamp}. We visualize the dynamic part. With integer time frames as input, baselines produce unstable geometry structures with distorted windows and wheels at fractional timestep, our method maintains the geometric coherence and local geometric structures of the dynamic components.}
    \label{exp_reg}
  \end{minipage}
  \hfill %
  \begin{minipage}[t]{7.cm}
  \vspace{0pt}
    \footnotesize
    \centering
    \renewcommand{\arraystretch}{1.1}
    \setlength{\tabcolsep}{4pt}
    
    \captionof{table}{\textbf{World Coordinate 4D Reconstruction.} We report performance on PO and TUM-Dynamics after global median scaling. All the experiments of methods are carried out using the code and pre-trained model from their official GitHub. The best results are marked in bold.}
    \label{tab:reconstruction}
    \scalebox{0.88}{
      \begin{tabular}{@{}clllll@{}}
        \toprule
        \textbf{} & \multicolumn{1}{c}{\textbf{}} & \multicolumn{2}{c}{Point Odyssey} & \multicolumn{2}{c}{TUM-Dynamics} \\ \cmidrule(l){3-4} \cmidrule(l){5-6} 
        Category & \multicolumn{1}{c}{Method} & APD$\uparrow$ & EPE$\downarrow$ & APD$\uparrow$ & EPE$\downarrow$ \\ 
        \midrule
        \multirow{3}{*}{\begin{tabular}[c]{@{}c@{}}w/ Global \\ Align.\end{tabular}} & DUSt3R+GA & 43.90 & 0.609 & 70.49 & 0.315 \\
         & MASt3R+GA & 60.44 & 0.403 & 68.38 & 0.519 \\
         & MonST3R+GA & 72.31 & 0.263 & 63.87 & 0.343 \\ \midrule
        \multirow{7}{*}{Feedforward} & DUSt3R & 45.79 & 0.639 & 72.26 & 0.289 \\
         & MASt3R & 56.90 & 0.464 & 66.22 & 0.551 \\
         & MonST3R & 68.25 & 0.304 & 61.38 & 0.365 \\
         & POMATO & 66.50 & 0.385 & 49.80 & 0.509 \\
         & St4RTrack & 78.73 & 0.241 & 83.42 & 0.185 \\
         & V-DPM & 69.05 & 0.342 & 75.92 & 0.273  \\ \cmidrule{2-6}
         & \textbf{Uni4R (Ours)} & \textbf{80.71}  & \textbf{0.222} & \textbf{84.36} & \textbf{0.181} \\ \bottomrule
      \end{tabular}
    }
  \end{minipage}

\end{table}

\textbf{Kinematics-Aware Benchmark.} Existing point tracking protocols mainly evaluate at integer timestamps, lacking continuous-time metrics. To bridge this gap, we introduce a kinematics-aware benchmark to assess whether continuous trajectories maintain kinematic coherence. Constructed from Kubric dynamic scenes~\cite{greff2022kubric}, our benchmark provides dense GT at both integer and fractional timestamps. We evaluate kinematic coherence via two aspects, temporal smoothness Mean Acceleration Magnitude~\cite{flash1985coordination} (MAM) and spatial structural stability based on integer and fractional timestamps, including Edge Length Variance (ELV), Variance of Local Pairwise Distances~\cite{sorkine2007rigid} (VLPD) and APD. Moreover, we also compute GT's MAM, ELV, and VLPD to evaluate absolute deviation $\Delta$ between each method and GT. More details are provided in Appendix.

\section{Experiments} \label{exps}

\textbf{Datasets and Training.} We train our model on 8 H800 GPUs using three datasets containing dynamic scene and camera motions: Point Odyssey~\cite{zheng2023pointodyssey}, Dynamic Replica~\cite{karaev2023dynamicstereo}, and Kubric~\cite{greff2022kubric}, which provide ground-truth at integer timestamps. During evaluation, we utilize an ODE's integration step size of $0.1$ between adjacent integer frames for our models. Detailed studies regarding the impact of step size on accuracy and inference time, training details, are available in Appendix.

\subsection{Point Tracking in World Coordinates} \label{exp_pt}

\textbf{Benchmark.} Following St4RTrack~\cite{feng2025st4rtrack}, we employ their proposed benchmark for point tracking in world coordinates. Specifically, the benchmark leverages two real-world datasets, including Aerial Digital Twin (ADT)~\cite{pan2023aria} and Panoptic Studio (PStudio)~\cite{7410738}, and two additional synthetic test sets from Point Odyssey (PO) and Dynamic Replica (DR). 

\begin{figure}[t]%
\centering
\includegraphics[width=0.99\textwidth]{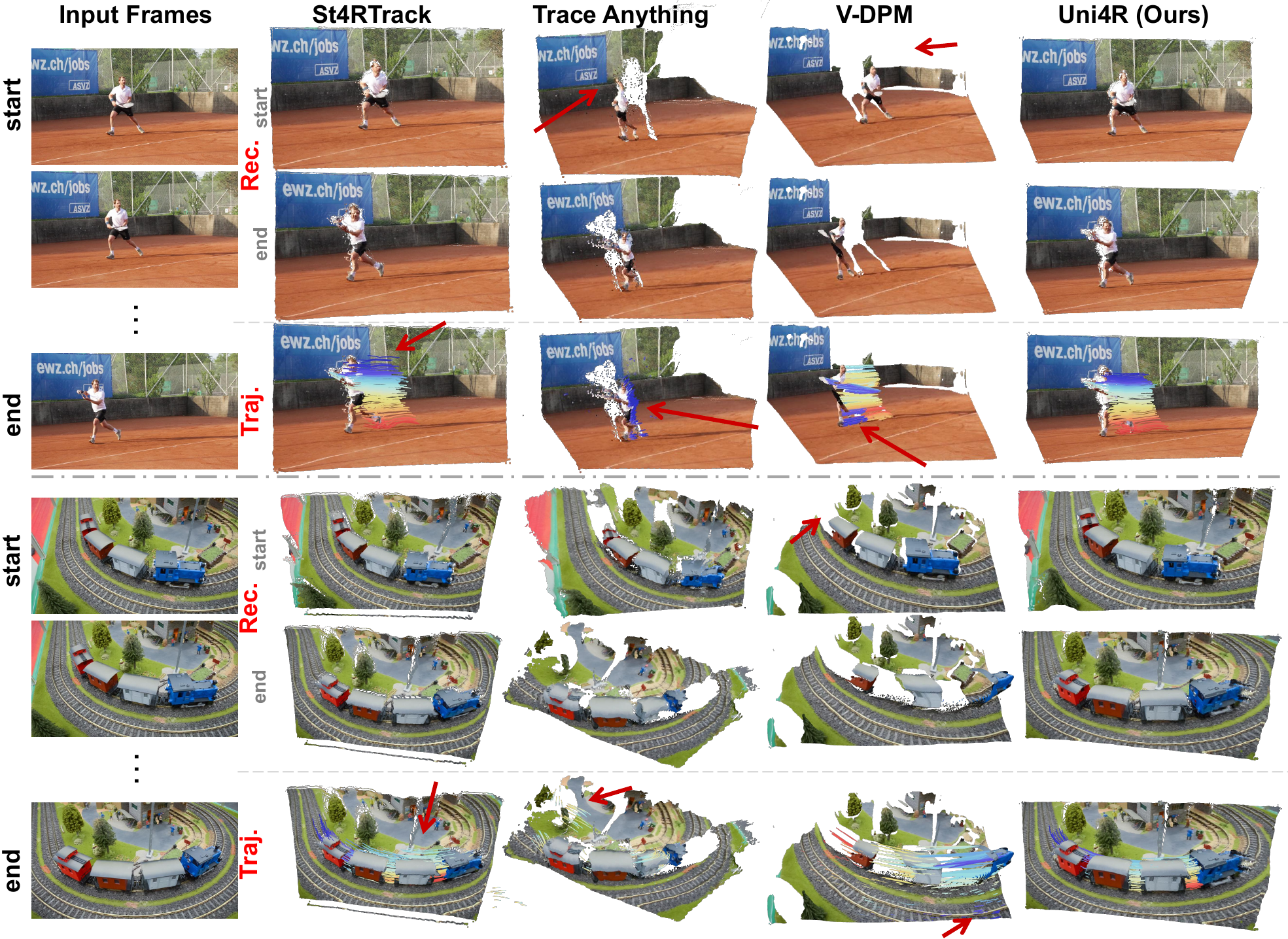}
\caption{\textbf{Qualitative Results of 4D Reconstruction and Point Tracking on real-world dataset DAVIS}~\cite{perazzi2016benchmark}. We visualize our Uni4R and baselines, including St4RTrack~\cite{feng2025st4rtrack} (we apply their test-time adaptation), Trace Anything~\cite{liu2025trace}, V-DPM~\cite{sucar2026v}. Our Uni4R effectively captures the geometric structure and motion trajectories in complex motion scenarios, whereas baselines produce incomplete geometry reconstruction (e.g., background and the moving train), and trajectories drift.}\label{exp_figs}
\vspace{-0.5cm}
\end{figure}

\textbf{Metrics.} Following St4RTrack~\cite{feng2025st4rtrack} and TAPVid-3D protocol~\cite{koppula2024tapvid}, we use Average percent of Points within Delta (APD) for evaluation. Following St4RTrack~\cite{feng2025st4rtrack}, we first align the predicted point trajectories with the ground truth through normalizing them with their global median. Then we compute the prediction error and we measure the percentage of points whose error falls below thresholds $\delta_{3D}\in\{0.1m,0.3m,0.5m,1.0m\}$ over the first 64 frames. Let $\hat{\mathbf{P}}_t^i$ denote the $i$-th point at time $t$, and $\mathbf{P}_t^i$ is its corresponding ground-truth location. Formally, $\mathrm{APD}_{3\mathrm{D}} \equiv \sum_{i,t} \mathds{1} \left( \left\| \hat{\mathbf{P}}_t^i - \mathbf{P}_t^i \right\| < \delta_{3D} \right)$, where $\mathds{1}$ denotes indicator function and $\left\|\cdot\right\|$ is Euclidean norm.

\textbf{Baselines.} Since we perform 3D tracking in world coordinates following St4RTrack, we first compare against the camera coordinate method SpatialTracker~\cite{xiao2024spatialtracker} and a dynamic 3D reconstruction method MonST3R~\cite{zhang2024monst3r}. In addition, we compare two combinational baselines in world coordinate 3D tracking, which applies Procrustes alignment and RANSAC to the camera coordinate 3D tracks predicted by SpatialTracker to offset the camera motion, and the second is the model which leverages the camera poses predicted by the MonST3R~\cite{fischler1981random} to compensate for camera motion~\cite{feng2025st4rtrack}. Moreover, we also compare a dynamic 3D tracking method POMATO~\cite{zhang2025pomato}, and the recent dynamic 3D tracking method TraceAnything~\cite{liu2025trace}, St4RTrack~\cite{feng2025st4rtrack}, and V-DPM~\cite{sucar2026v}.

\textbf{Results.} As shown in Tab.~\ref{tab:3dtracking}, we achieve SOTA performance in all test datasets and metrics. The performance gap highlights the lack of explicit kinematic priors in baselines. Thus their predictions at integer timestamps remain inherently inaccurate under complex motion scenarios. As shown in Fig.~\ref{exp_figs}, baselines suffer from trajectory drift. In contrast, our framework resolves this by synergizing OT and ODE to learn a continuous velocity field. The continuous velocity field acts as a robust kinematic prior. By leveraging this prior, Uni4R yields significantly more accurate spatial-temporal correspondences to predict point tracking. Besides, in Tab.~\ref{tab:efficiency}. our Uni4R achieves a highly efficient inference speed (1.57s), trailing only St4RTrack. Although Uni4R has slightly higher VRAM footprint compared to Trace Anything is heavily outweighed by a massive performance margin.

\begin{table}[t]
  \centering
  
  %
  \begin{minipage}[t]{0.32\textwidth}
    \centering
    \caption{\textbf{Efficiency comparison on 80-frame input}. We test on the same H800 GPU. Our Uni4R employs 10 integration steps between adjacent integer frames.}
    \label{tab:efficiency}

    \resizebox{1\linewidth}{!}{
      \begin{tabular}{@{} l c c @{}} 
        \toprule
        Methods & Time (s) & VRAM (GB) \\
        \midrule
        St4RTrack & 1.34 & 24.06 \\
        Trace Anything & 4.69 & 35.71 \\
        V-DPM & 194.14 & 75.27 \\
        \midrule
        \textbf{Uni4R (Ours)} & 1.57 & 43.67 \\
        \bottomrule
      \end{tabular}
    } %
  \end{minipage}
  \hfill %
  %
  \begin{minipage}[t]{0.66\textwidth}
    \centering
    \caption{\textbf{Kinematic Coherence in Continuous Time.} We report the performance on kinematics-aware Benchmark after global median alignment.The best results are marked in bold.}
    \label{tab:physical}
    
    \resizebox{1\linewidth}{!}{
      \begin{tabular}{@{}lcccc@{}}
        \toprule
        \multirow{2}{*}{Methods} & Temporal Kinematic & \multicolumn{3}{c}{Spatial Topology} \\ 
        \cmidrule(lr){2-2} \cmidrule(lr){3-5} 
         & MAM $\downarrow$ / $\Delta\downarrow$ & \makecell{$\text{APD}_\text{3D}$ $\uparrow$} & \makecell{ELV $\downarrow$  / $\Delta\downarrow$} & VLPD $\downarrow$  / $\Delta\downarrow$ \\ \midrule
         Ground Truth (GT) & 6.99 / - & - & 10.2316 / - & 0.0003 / - \\ \midrule 
        Baseline w/ B-Spline & 15.93 / 8.94 & 75.04 & 10.3107 / 0.0791 & 0.0274 / 0.0271   \\ 
        Baseline w/ LVP + B-Spline & 13.75 / 6.76 & 76.37 & 10.2989 / 0.0673 & 0.0146 / 0.0143  \\ 
        Baseline w/ LVP + ODE & 13.64 / 6.65 & 76.89 & 10.2970 / 0.0654 & 0.0115 / 0.0112 \\ 
        TraceAnything & 7.89 / 0.90 & 77.90 & 10.2553 / 0.0237 & 0.0015 / 0.0012  \\ \midrule 
        \textbf{Uni4R (Ours)} & \textbf{7.27} / \textbf{0.28} & \textbf{79.53} & \textbf{10.2437} / \textbf{0.0121} & \textbf{0.0004} / \textbf{0.0001}   \\ \bottomrule
      \end{tabular}
    }
  \end{minipage}

\end{table}

\subsection{4D Reconstruction} \label{exp_rec}

\textbf{Benchmark.} Following St4RTrack~\cite{feng2025st4rtrack}, we evaluate the quality of 4D reconstruction on synthetic dataset Point Odyssey, and real-world dataset TUM-Dynamics.

\textbf{Metrics.} Following St4RTrack, we compare the reconstructed 3D point clouds to the ground truth using the Average percent of Points within Distance (APD) and End-Point Error (EPE) metrics. Before evaluation, we filter out ambiguous floating points from the ground-truth data and align the point clouds for each sequence using the median scale.

\textbf{Baselines.} We compare our Uni4R with baselines, including MonST3R, MASt3R, DUSt3R, POMATO, recent methods St4RTrack, and V-DPM. Following St4RTrack, for the global align (GA), we apply global alignment on baselines; For the feedforward baselines, we construct image pairs of a video that align all frames to a common anchor frame.

\textbf{Results.} As shown in Tab.~\ref{tab:reconstruction}, we achieve SOTA performance in all test datasets and metrics. Because baselines ignore constructing or using kinematic priors, thus their predictions at integer timestamps remain fragmented geometry and topological distortions, as shown in Fig.~\ref{exp_figs}. In contrast, our Uni4R constructs and leverages kinematic prior from continuous velocity field, Uni4R effectively captures the geometric structure in complex motion scenarios.

\subsection{Continuous-Time Kinematic Coherence} \label{exp_phys}

\textbf{Benchmark.} We employ our kinematics-aware benchmark to evaluate our Uni4R and some baselines on the Kinematic Coherence in continuous time. The benchmark includes 100 diverse dynamic scene. Each dynamic scenes has 24 integer frames. And we render 24 integer and 230 fractional timestamps' depth, masks, camera poses and ground-truth point trajectories for each dynamic scene.

\textbf{Metrics.} As we introduce in Sec.~\ref{ic_loss}, we employ MAM, ELV, VLPD and APD to evaluate the kinematics coherence in continuous time, and also report the MAM, ELV, VLPD absolute deviation $\Delta$ between each method and GT. We evaluate on both integer and fractional timestamps.

\textbf{Baselines.} Since most recent works focus on integer timestamps, only few works perform on fractional timestamps, e.g., TraceAnything~\cite{liu2025trace}. Therefore, we compare our Uni4R with TraceAnything. Besides, we also construct a series baseline, which replaces FMGD with two decoder and two DPT. One decoder and DPT predicts the reconstructed point maps, while another predicts velocities on integer timestamps. Then the velocities will be interpolated via B-Spline (Baseline w/ B-Spline) on fractional timestamps; or employ LVP to predict control points to control the velocity (Baseline w/ LVP + B-Spline) on fractional timestamps; or employ MLPs and ODE to predict velocities (Baseline w/ LVP + ODE) on fractional timestamps.

\textbf{Results.} As shown in Tab.~\ref{tab:physical}, we achieve SOTA performance in all metrics, proving ours superior kinematic coherence. Our Uni4R achieves the closest overall agreement with GT among the compared baselines. Besides, our low MAM verifies smooth continuous trajectories, while low ELV and VLPD, and high APD confirm the preservation of local geometric structures. As shown in Fig.~\ref{exp_reg}, baselines lack kinematic priors to constrain continuous geometry deformations, causing severe distortions like warped wheels and stretched windows. In contrast, Uni4R learns a continuous velocity field that acts as a robust kinematic prior. This kinematic prior preserves local geometric structures, ensuring topologically consistent geometric structures at any fractional timestamp.

\subsection{Ablation Study} 
\textbf{Basic Ablations:} To validate the effect of our proposed components, we progressively remove the core components within FMGD of Uni4R in Tab.~\ref{tab:ablation1}. Starting from Uni4R, removing the continuous flow matching loss (w/o $\mathcal{L}_{CFM}$) leads to a notable performance drop in all tasks due to the lack of continuous velocity regularization. Further removing the FM-guided Velocity Features (w/o FVF) leads that causes the model to lose its kinematic inductive bias, causing a continued decline in all tasks accuracy, especially  kinematic coherence. Next, we remove integral-consistency training strategy (w/o ICT) means remove the end-to-end velocity supervision, resulting in a performance degradation. It confirms that simply applying an ODE is insufficient because of ill-posed problem, leading to low performance compared to previous model. Finally, we omit the Local Velocity Prediction module (w/o LVP) and replace it using simple cross-attention, which cannot accurate captures geometry and motion information to predict precise velocity, yielding the lowest overall performance. Besides, we replace Uni4R’s anchor-based ODE integration by global-based ODE (model G), which leads to a performance drop in all tasks due to the accumulation of integration errors. It validates that we discuss in Sec. \ref{method_ode}, anchor-based ODE mitigates the accumulation of integration errors.

\begin{wraptable}{r}{8.7cm}
\footnotesize
\setlength{\tabcolsep}{6pt}
\caption{\textbf{Ablation Experiments} on 4D Reconstruction and Point Tracking (report $\text{APD}_\text{3D}$ $\uparrow$), and continuous-time kinematic coherence (named KC, we report ELV $\downarrow$). \textbf{Top}: we present the baselines using continuous modeling variant. \textbf{Middle}: We compare models which cumulatively (from C to F) remove core components within FMGD of Uni4R. \textbf{Below}: We replace Uni4R's anchor-based ODE integration by global-based ODE, which directly integrate velocity from time 0 to last.}
\vspace{-5pt}
\label{tab:ablation1}
\scalebox{0.76}{\begin{tabular}{@{}lccccccc@{}}
\toprule
\multirow{2}{*}{Methods} & \multicolumn{4}{c}{Point Tracking} & \multicolumn{2}{c}{Reconstruction} & \multirow{2}{*}{KC} \\ 
\cmidrule(lr){2-5} \cmidrule(lr){6-7} 
& PO & DR & ADT & PStudio & PO & TUM & \\ 
\midrule
A. Baseline & 74.31 & 74.19 & 79.83 & 74.66 & 77.94 & 81.05 & / \\
B. Baseline w/ LVP + B-Spline & 74.99 & 74.41 & 81.26 & 76.21 & 78.38 & 82.14 & 10.2989  \\ \midrule

\textbf{Uni4R (Ours)} & \textbf{81.59} & \textbf{76.05} & \textbf{86.55} & \textbf{79.43} & \textbf{80.71} & \textbf{84.36} & \textbf{10.2437} \\
C. w/o $\mathcal{L}_{CFM}$ & 80.16 & 75.71  & 85.43 & 78.94 & 79.88 & 84.11 & 10.2569  \\
D. w/o FVF  & 79.36 & 75.43 & 83.35 & 77.12 & 78.95 & 83.72 & 10.2728  \\
E. w/o ICT & 75.27 & 74.65 & 81.50 & 76.39 & 78.52 & 82.93 & 10.2951  \\ 
F. w/o LVP  & 74.68 & 74.48 & 80.63 & 75.07 & 78.33 & 81.76 & 10.3045  \\  \midrule
G. Global-based ODE & 79.10 & 75.57 & 85.38 & 79.02 & 78.73 & 83.76 & 10.2819 \\ 
\bottomrule
\end{tabular}}
\vspace{-2pt}
\end{wraptable}

\textbf{Comparison with Continuous Modeling Variant:} In Tab.~\ref{tab:ablation1} and Fig.~\ref{exp_reg}, we further evaluate our Uni4R against continuous modeling variants based on Baseline. The Baseline replaces FMGD with two decoder and two DPT. One decoder and DPT predict reconstructed point maps, while another predicts velocities on integer timestamps. Based on Baseline, model B which first employ LVP to predict control points then applying B-spline yields marginal improvements in both two tasks. But there exists a large performance gap compared to our Uni4R and model E (model E means to directly apply ODE without ICT). Because model B focuses on geometric curve fitting instead of leveraging kinematic priors for dynamic modeling.

\begin{wraptable}{r}{7.5cm}
\footnotesize
\setlength{\tabcolsep}{6pt}
\caption{\textbf{Ablation Experiments} on the Mutual Enhancement of Reconstruction and Tracking. We compare the models whether train velocity branch (vel.) or train reconstruction branch (rec.). Each model loads the same pre-trained weights before training.}
\vspace{-5pt}
\label{tab:ablation3}
\scalebox{0.79}{\begin{tabular}{@{}lcccccccc@{}} 
\toprule
\multirow{2}{*}{Methods} & \multicolumn{2}{c}{Train} & \multicolumn{4}{c}{Point Tracking} & \multicolumn{2}{c}{Reconstruction} \\ \cmidrule(lr){2-3} \cmidrule(l){4-7} \cmidrule(l){8-9} 
 & Vel. & Rec. & PO & DR & ADT & PStudio & PO & TUM \\ \midrule
M1 & \Checkmark & \XSolidBrush & 79.74 & 75.18 & 84.89 & 78.67 & 78.89 & 83.01 \\ 
M2 & \XSolidBrush & \Checkmark & 78.72 & 74.05 & 82.94 & 78.01 & 79.78 & 83.98\\
\textbf{Ours} & \Checkmark & \Checkmark & \textbf{81.59} & \textbf{76.05} & \textbf{86.55} & \textbf{79.43} & \textbf{80.71} & \textbf{84.36} \\ \bottomrule
\end{tabular}}
\vspace{-10pt}
\end{wraptable}

\textbf{Mutual Enhancement of Reconstruction and Tracking:} To investigate the mutual benefit between 4D reconstruction and point tracking within Uni4R, we evaluate models under different training strategies (with the same pre-trained models) in Tab.~\ref{tab:ablation3}. Training only the velocity branch (model M1) or the reconstruction branch (model M2) leads to lower performance than our Uni4R both in two tasks. Because separate optimization breaks the inherent coupling between geometry and motion. In contrast, joint training (our Uni4R) unlocks the full potential of learned continuous velocity field. Specifically, the continuous velocity filed acts as a joint kinematic prior. The accurate point tracking provides kinematic guidance for geometric deformation, while high-quality 4D reconstruction regularizes the trajectory space. This kinematics-aware and unified framework allows the two tasks to mutually enhance each other, yielding superior performance in both two tasks.

\section{Conclusion}

In this paper, we propose Uni4R, a novel framework that unifies 4D reconstruction and point tracking. By synergizing Optimal Transport (OT) and Ordinary Differential Equations (ODE), our method learns a continuous velocity field that serves as a robust kinematic prior, ensuring kinematic coherence at any arbitrary timestamp. A key component is the Flow Matching Guided Decoder (FMGD), which effectively leverages FM theory to decode these continuous velocities. Furthermore, our integral-consistency training strategy addresses the absence of fractional ground truth velocities, enabling end-to-end supervision directly from integer frames. Our method achieves state-of-the-art performance across two tasks, and our new kinematics-aware benchmark.

\paragraph{Acknowledgments}
This work is supported by the Science and Technology Development Fund of Macau  Project (grant No. 0096/2023/RIA2, 0044/2024/AGJ, 0140/2024/AGJ, 0084/2024/RIB2), and is partially supported by the AI Supercomputing Center of Macau University of Science and Technology (MUST). We also acknowledge the technical support provided by the Information Technology Development Office of MUST.

\bibliographystyle{Ref}  
\small
\bibliography{ref}

\begin{thebibliography}{69}
\providecommand{\natexlab}[1]{#1}
\providecommand{\url}[1]{\texttt{#1}}
\expandafter\ifx\csname urlstyle\endcsname\relax
  \providecommand{\doi}[1]{doi: #1}\else
  \providecommand{\doi}{doi: \begingroup \urlstyle{rm}\Url}\fi

\bibitem[Benamou and Brenier(2000)]{benamou2000computational}
Benamou , J.-D. \& Brenier , Y. (2000)
\newblock A computational fluid mechanics solution to the monge-kantorovich mass transfer problem.
\newblock \emph{Numerische Mathematik} {\bfseries 84}\penalty0 (3):\penalty0 375--393.

\bibitem[Chen et~al.(2018{\natexlab{a}})Chen, Li, Chen, Wang, Pu, and Duke]{chen2018continuous}
Chen , C., Li~, C., Chen , L., Wang , W., Pu~, Y., \& Duke , L.~C. (2018.
\newblock {\natexlab{a}}) Continuous-time flows for efficient inference and density estimation. In
\newblock \emph{International Conference on Machine Learning}
\newblock pages 824--833. PMLR.

\bibitem[Chen et~al.(2018{\natexlab{b}})Chen, Rubanova, Bettencourt, and Duvenaud]{chen2018neural}
Chen , R.~T., Rubanova , Y., Bettencourt , J., \& Duvenaud , D.~K. (2018.
\newblock {\natexlab{b}})
\newblock Neural ordinary differential equations.
\newblock \emph{Advances in neural information processing systems} {\bfseries 31}.

\bibitem[Dao et~al.(2022)Dao, Fu, Ermon, Rudra, and R{\'e}]{dao2022flashattention}
Dao , T., Fu~, D., Ermon , S., Rudra , A., \& R{\'e}~, C. (2022)
\newblock Flashattention: Fast and memory-efficient exact attention with io-awareness.
\newblock \emph{Advances in neural information processing systems} {\bfseries 35}:\penalty0 16344--16359.

\bibitem[Dupont et~al.(2019)Dupont, Doucet, and Teh]{dupont2019augmented}
Dupont , E., Doucet , A., \& Teh , Y.~W. (2019)
\newblock Augmented neural odes.
\newblock \emph{Advances in neural information processing systems} {\bfseries 32}.

\bibitem[Farin(2002)]{farin2002curves}
Farin , G.~E. (2002)
\newblock \emph{Curves and surfaces for CAGD: a practical guide}.
\newblock \emph{Curves and surfaces for CAGD: a practical guide}: Morgan Kaufmann.

\bibitem[Feng et~al.(2025)Feng, Zhang, Wang, Ye, Yu, Black, Darrell, and Kanazawa]{feng2025st4rtrack}
Feng , H., Zhang , J., Wang , Q., Ye~, Y., Yu~, P., Black , M.~J., Darrell , T., \& Kanazawa , A. (2025)
\newblock St4rtrack: Simultaneous 4d reconstruction and tracking in the world. In
\newblock \emph{Proceedings of the IEEE/CVF International Conference on Computer Vision}
\newblock pages 8503--8513.

\bibitem[Fischler and Bolles(1981)]{fischler1981random}
Fischler , M.~A. \& Bolles , R.~C. (1981)
\newblock Random sample consensus: a paradigm for model fitting with applications to image analysis and automated cartography.
\newblock \emph{Communications of the ACM} {\bfseries 24}\penalty0 (6):\penalty0 381--395.

\bibitem[Flash and Hogan(1985)]{flash1985coordination}
Flash , T. \& Hogan , N. (1985)
\newblock The coordination of arm movements: an experimentally confirmed mathematical model.
\newblock \emph{Journal of neuroscience} {\bfseries 5}\penalty0 (7):\penalty0 1688--1703.

\bibitem[Grathwohl et~al.(2018)Grathwohl, Chen, Bettencourt, Sutskever, and Duvenaud]{grathwohl2018ffjord}
Grathwohl , W., Chen , R.~T., Bettencourt , J., Sutskever , I., \& Duvenaud , D. (2018)
\newblock Ffjord: Free-form continuous dynamics for scalable reversible generative models.
\newblock \emph{arXiv preprint arXiv:1810.01367}

\bibitem[Greff et~al.(2022)Greff, Belletti, Beyer, Doersch, Du, Duckworth, Fleet, Gnanapragasam, Golemo, Herrmann, et~al.]{greff2022kubric}
Greff , K., Belletti , F., Beyer , L., Doersch , C., Du~, Y., Duckworth , D., Fleet , D.~J., Gnanapragasam , D., Golemo , F., Herrmann , C., \& others  (2022)
\newblock Kubric: A scalable dataset generator. In
\newblock \emph{Proceedings of the IEEE/CVF conference on computer vision and pattern recognition}
\newblock pages 3749--3761.

\bibitem[Hunyuan3D et~al.(2025)Hunyuan3D, Zhang, Guo, Liu, Yan, Shi, Huang, Yu, Li, Wang, et~al.]{hunyuan3d2025hunyuan3d}
Hunyuan3D , T., Zhang , B., Guo , C., Liu , H., Yan , H., Shi , H., Huang , J., Yu~, J., Li~, K., Wang , P., \& others  (2025)
\newblock Hunyuan3d-omni: A unified framework for controllable generation of 3d assets.
\newblock \emph{arXiv preprint arXiv:2509.21245}

\bibitem[Joo et~al.(2015)Joo, Liu, Tan, Gui, Nabbe, Matthews, Kanade, Nobuhara, and Sheikh]{7410738}
Joo , H., Liu , H., Tan , L., Gui , L., Nabbe , B., Matthews , I., Kanade , T., Nobuhara , S., \& Sheikh , Y. (2015)
\newblock Panoptic studio: A massively multiview system for social motion capture. In
\newblock \emph{2015 IEEE International Conference on Computer Vision (ICCV)}
\newblock pages 3334--3342.

\bibitem[Karaev et~al.(2023)Karaev, Rocco, Graham, Neverova, Vedaldi, and Rupprecht]{karaev2023dynamicstereo}
Karaev , N., Rocco , I., Graham , B., Neverova , N., Vedaldi , A., \& Rupprecht , C. (2023)
\newblock Dynamicstereo: Consistent dynamic depth from stereo videos. In
\newblock \emph{Proceedings of the IEEE/CVF Conference on Computer Vision and Pattern Recognition}
\newblock pages 13229--13239.

\bibitem[Karaev et~al.(2024)Karaev, Rocco, Graham, Neverova, Vedaldi, and Rupprecht]{karaev2024cotracker}
Karaev , N., Rocco , I., Graham , B., Neverova , N., Vedaldi , A., \& Rupprecht , C. (2024)
\newblock Cotracker: It is better to track together. In
\newblock \emph{European conference on computer vision}
\newblock pages 18--35. Springer.

\bibitem[Karaev et~al.(2025)Karaev, Makarov, Wang, Neverova, Vedaldi, and Rupprecht]{karaev2025cotracker3}
Karaev , N., Makarov , Y., Wang , J., Neverova , N., Vedaldi , A., \& Rupprecht , C. (2025)
\newblock Cotracker3: Simpler and better point tracking by pseudo-labelling real videos. In
\newblock \emph{Proceedings of the IEEE/CVF International Conference on Computer Vision}
\newblock pages 6013--6022.

\bibitem[Kerbl et~al.(2023)Kerbl, Kopanas, Leimk{\"u}hler, Drettakis, et~al.]{kerbl20233d}
Kerbl , B., Kopanas , G., Leimk{\"u}hler , T., Drettakis , G., \& others  (2023)
\newblock 3d gaussian splatting for real-time radiance field rendering.
\newblock \emph{ACM Trans. Graph.} {\bfseries 42}\penalty0 (4):\penalty0 139--1.

\bibitem[Koppula et~al.(2024)Koppula, Rocco, Yang, Heyward, Carreira, Zisserman, Brostow, and Doersch]{koppula2024tapvid}
Koppula , S., Rocco , I., Yang , Y., Heyward , J., Carreira , J., Zisserman , A., Brostow , G., \& Doersch , C. (2024)
\newblock Tapvid-3d: A benchmark for tracking any point in 3d.
\newblock \emph{Advances in Neural Information Processing Systems} {\bfseries 37}:\penalty0 82149--82165.

\bibitem[Lipman et~al.(2022)Lipman, Chen, Ben-Hamu, Nickel, and Le]{lipman2022flow}
Lipman , Y., Chen , R.~T., Ben-Hamu , H., Nickel , M., \& Le~, M. (2022)
\newblock Flow matching for generative modeling.
\newblock \emph{arXiv preprint arXiv:2210.02747}

\bibitem[Liu et~al.(2023)Liu, Wu, Van~Hoorick, Tokmakov, Zakharov, and Vondrick]{liu2023zero}
Liu , R., Wu~, R., Van~Hoorick , B., Tokmakov , P., Zakharov , S., \& Vondrick , C. (2023)
\newblock Zero-1-to-3: Zero-shot one image to 3d object. In
\newblock \emph{Proceedings of the IEEE/CVF international conference on computer vision}
\newblock pages 9298--9309.

\bibitem[Liu et~al.(2022)Liu, Gong, and Liu]{liu2022flow}
Liu , X., Gong , C., \& Liu , Q. (2022)
\newblock Flow straight and fast: Learning to generate and transfer data with rectified flow.
\newblock \emph{arXiv preprint arXiv:2209.03003}

\bibitem[Liu et~al.(2025)Liu, Xiao, Chen, Feng, Tai, Tang, and Kang]{liu2025trace}
Liu , X., Xiao , Y., Chen , D.~Y., Feng , J., Tai , Y.-W., Tang , C.-K., \& Kang , B. (2025)
\newblock Trace anything: Representing any video in 4d via trajectory fields.
\newblock \emph{arXiv preprint arXiv:2510.13802}

\bibitem[Long et~al.(2024)Long, Guo, Lin, Liu, Dou, Liu, Ma, Zhang, Habermann, Theobalt, et~al.]{long2024wonder3d}
Long , X., Guo , Y.-C., Lin , C., Liu , Y., Dou , Z., Liu , L., Ma~, Y., Zhang , S.-H., Habermann , M., Theobalt , C., \& others  (2024)
\newblock Wonder3d: Single image to 3d using cross-domain diffusion. In
\newblock \emph{Proceedings of the IEEE/CVF Conference on Computer Vision and Pattern Recognition}
\newblock pages 9970--9980.

\bibitem[Lu et~al.(2026)Lu, Xu, Hu, Zhu, Zhao, Yeung, Shan, and Liu]{lu2026track4world}
Lu~, J., Xu~, J., Hu~, W., Zhu , R., Zhao , C., Yeung , S.-K., Shan , Y., \& Liu , Y. (2026)
\newblock Track4world: Feedforward world-centric dense 3d tracking of all pixels. In
\newblock \emph{European Conference on Computer Vision}
\newblock pages 184--204. Springer.

\bibitem[Lyu et~al.(2026)Lyu, Geng, Dharmarajan, Zhang, Alzayer, Wu, and Wu]{lyu2026choreographing}
Lyu , Y., Geng , C., Dharmarajan , K., Zhang , Y., Alzayer , H., Wu~, S., \& Wu~, J. (2026)
\newblock Choreographing a world of dynamic objects.
\newblock \emph{arXiv preprint arXiv:2601.04194}

\bibitem[Melas-Kyriazi et~al.(2023)Melas-Kyriazi, Rupprecht, and Vedaldi]{melas2023pc}
Melas-Kyriazi , L., Rupprecht , C., \& Vedaldi , A. (2023)
\newblock Pc 2: Projection-conditioned point cloud diffusion for single-image 3d reconstruction. In
\newblock \emph{2023 IEEE/CVF Conference on Computer Vision and Pattern Recognition (CVPR)}
\newblock pages 12923--12932. IEEE.

\bibitem[Norcliffe et~al.(2020)Norcliffe, Bodnar, Day, Simidjievski, and Li{\`o}]{norcliffe2020second}
Norcliffe , A., Bodnar , C., Day , B., Simidjievski , N., \& Li{\`o} , P. (2020)
\newblock On second order behaviour in augmented neural odes.
\newblock \emph{Advances in neural information processing systems} {\bfseries 33}:\penalty0 5911--5921.

\bibitem[Pan et~al.(2023)Pan, Charron, Yang, Peters, Whelan, Kong, Parkhi, Newcombe, and Ren]{pan2023aria}
Pan , X., Charron , N., Yang , Y., Peters , S., Whelan , T., Kong , C., Parkhi , O., Newcombe , R., \& Ren , Y.~C. (2023)
\newblock Aria digital twin: A new benchmark dataset for egocentric 3d machine perception. In
\newblock \emph{Proceedings of the IEEE/CVF International Conference on Computer Vision}
\newblock pages 20133--20143.

\bibitem[Perazzi et~al.(2016)Perazzi, Pont-Tuset, McWilliams, Van~Gool, Gross, and Sorkine-Hornung]{perazzi2016benchmark}
Perazzi , F., Pont-Tuset , J., McWilliams , B., Van~Gool , L., Gross , M., \& Sorkine-Hornung , A. (2016)
\newblock A benchmark dataset and evaluation methodology for video object segmentation. In
\newblock \emph{Proceedings of the IEEE conference on computer vision and pattern recognition}
\newblock pages 724--732.

\bibitem[Poole et~al.(2022)Poole, Jain, Barron, and Mildenhall]{poole2022dreamfusion}
Poole , B., Jain , A., Barron , J.~T., \& Mildenhall , B. (2022)
\newblock Dreamfusion: Text-to-3d using 2d diffusion.
\newblock \emph{arXiv preprint arXiv:2209.14988}

\bibitem[Qian et~al.(2026)Qian, Zhang, Wu, and Cremers]{qian2026flow4r}
Qian , S., Zhang , G., Wu~, S., \& Cremers , D.
\newblock 2026.

\bibitem[Ren et~al.(2024)Ren, Xie, Mirzaei, Liang, Zeng, Kreis, Liu, Torralba, Fidler, Kim, et~al.]{ren2024l4gm}
Ren , J., Xie , K., Mirzaei , A., Liang , H., Zeng , X., Kreis , K., Liu , Z., Torralba , A., Fidler , S., Kim , S.~W., \& others  (2024)
\newblock L4gm: Large 4d gaussian reconstruction model.
\newblock \emph{Advances in Neural Information Processing Systems} {\bfseries 37}:\penalty0 56828--56858.

\bibitem[Shi et~al.(2023)Shi, Chen, Zhang, Liu, Xu, Wei, Chen, Zeng, and Su]{shi2023zero123++}
Shi , R., Chen , H., Zhang , Z., Liu , M., Xu~, C., Wei , X., Chen , L., Zeng , C., \& Su~, H. (2023)
\newblock Zero123++: a single image to consistent multi-view diffusion base model.
\newblock \emph{arXiv preprint arXiv:2310.15110}

\bibitem[Singer et~al.(2022)Singer, Polyak, Hayes, Yin, An, Zhang, Hu, Yang, Ashual, Gafni, et~al.]{singer2022make}
Singer , U., Polyak , A., Hayes , T., Yin , X., An~, J., Zhang , S., Hu~, Q., Yang , H., Ashual , O., Gafni , O., \& others  (2022)
\newblock Make-a-video: Text-to-video generation without text-video data.
\newblock \emph{arXiv preprint arXiv:2209.14792}

\bibitem[Sorkine et~al.(2007)Sorkine, Alexa, et~al.]{sorkine2007rigid}
Sorkine , O., Alexa , M., \& others  (2007)
\newblock As-rigid-as-possible surface modeling. In
\newblock \emph{Symposium on Geometry processing}
\newblock \emph{4}, pp. \penalty0 109--116.

\bibitem[Sucar et~al.(2025)Sucar, Lai, Insafutdinov, and Vedaldi]{sucar2025dynamic}
Sucar , E., Lai , Z., Insafutdinov , E., \& Vedaldi , A. (2025)
\newblock Dynamic point maps: A versatile representation for dynamic 3d reconstruction. In
\newblock \emph{Proceedings of the IEEE/CVF International Conference on Computer Vision}
\newblock pages 7295--7305.

\bibitem[Sucar et~al.(2026)Sucar, Insafutdinov, Lai, and Vedaldi]{sucar2026v}
Sucar , E., Insafutdinov , E., Lai , Z., \& Vedaldi , A. (2026)
\newblock V-dpm: 4d video reconstruction with dynamic point maps.
\newblock \emph{arXiv preprint arXiv:2601.09499}

\bibitem[Tang et~al.(2024)Tang, Chen, Chen, Wang, Zeng, and Liu]{tang2024lgm}
Tang , J., Chen , Z., Chen , X., Wang , T., Zeng , G., \& Liu , Z. (2024)
\newblock Lgm: Large multi-view gaussian model for high-resolution 3d content creation. In
\newblock \emph{European Conference on Computer Vision}
\newblock pages 1--18. Springer.

\bibitem[Tong et~al.(2023)Tong, Fatras, Malkin, Huguet, Zhang, Rector-Brooks, Wolf, and Bengio]{tong2023improving}
Tong , A., Fatras , K., Malkin , N., Huguet , G., Zhang , Y., Rector-Brooks , J., Wolf , G., \& Bengio , Y. (2023)
\newblock Improving and generalizing flow-based generative models with minibatch optimal transport.
\newblock \emph{arXiv preprint arXiv:2302.00482}

\bibitem[Wan et~al.(2025)Wan, Wang, Ai, Wen, Mao, Xie, Chen, Yu, Zhao, Yang, et~al.]{wan2025wan}
Wan , T., Wang , A., Ai~, B., Wen , B., Mao , C., Xie , C.-W., Chen , D., Yu~, F., Zhao , H., Yang , J., \& others  (2025)
\newblock Wan: Open and advanced large-scale video generative models.
\newblock \emph{arXiv preprint arXiv:2503.20314}

\bibitem[Wang et~al.(2025{\natexlab{a}})Wang, Chen, Karaev, Vedaldi, Rupprecht, and Novotny]{wang2025vggt}
Wang , J., Chen , M., Karaev , N., Vedaldi , A., Rupprecht , C., \& Novotny , D. (2025.
\newblock {\natexlab{a}}) Vggt: Visual geometry grounded transformer. In
\newblock \emph{Proceedings of the Computer Vision and Pattern Recognition Conference}
\newblock pages 5294--5306.

\bibitem[Wang et~al.(2024)Wang, Leroy, Cabon, Chidlovskii, and Revaud]{wang2024dust3r}
Wang , S., Leroy , V., Cabon , Y., Chidlovskii , B., \& Revaud , J. (2024)
\newblock Dust3r: Geometric 3d vision made easy. In
\newblock \emph{Proceedings of the IEEE/CVF conference on computer vision and pattern recognition}
\newblock pages 20697--20709.

\bibitem[Wang et~al.(2025{\natexlab{b}})Wang, Zhou, Zhu, Chang, Zhou, Li, Chen, Pang, Shen, and He]{wang2025pi}
Wang , Y., Zhou , J., Zhu , H., Chang , W., Zhou , Y., Li~, Z., Chen , J., Pang , J., Shen , C., \& He~, T. (2025.
\newblock {\natexlab{b}})
\newblock $\pi^3$: Permutation-equivariant visual geometry learning.
\newblock \emph{arXiv preprint arXiv:2507.13347}

\bibitem[Wang et~al.(2026)Wang, Wang, Zhang, Yu, and Chen]{wang2026mcgs}
Wang , Y., Wang , W., Zhang , S., Yu~, X., \& Chen , Z. (2026)
\newblock Mcgs: Markov chain gaussian splatting for dynamic scenes reconstruction. In
\newblock \emph{Proceedings of the AAAI Conference on Artificial Intelligence}
\newblock \emph{40}, pp. \penalty0 10341--10348.

\bibitem[Wu et~al.(2024)Wu, Yu, Jiang, Cao, Wang, and Bai]{wu2024sc4d}
Wu~, Z., Yu~, C., Jiang , Y., Cao , C., Wang , F., \& Bai , X. (2024)
\newblock Sc4d: Sparse-controlled video-to-4d generation and motion transfer. In
\newblock \emph{European Conference on Computer Vision}
\newblock pages 361--379. Springer.

\bibitem[Xiao et~al.(2024)Xiao, Wang, Zhang, Xue, Peng, Shen, and Zhou]{xiao2024spatialtracker}
Xiao , Y., Wang , Q., Zhang , S., Xue , N., Peng , S., Shen , Y., \& Zhou , X. (2024)
\newblock Spatialtracker: Tracking any 2d pixels in 3d space. In
\newblock \emph{Proceedings of the IEEE/CVF Conference on Computer Vision and Pattern Recognition}
\newblock pages 20406--20417.

\bibitem[Xiao et~al.(2025)Xiao, Wang, Xue, Karaev, Makarov, Kang, Zhu, Bao, Shen, and Zhou]{xiao2025spatialtrackerv2}
Xiao , Y., Wang , J., Xue , N., Karaev , N., Makarov , Y., Kang , B., Zhu , X., Bao , H., Shen , Y., \& Zhou , X. (2025)
\newblock Spatialtrackerv2: Advancing 3d point tracking with explicit camera motion. In
\newblock \emph{Proceedings of the IEEE/CVF International Conference on Computer Vision}
\newblock pages 6726--6737.

\bibitem[Xie et~al.(2019)Xie, Yao, Sun, Zhou, and Zhang]{xie2019pix2vox}
Xie , H., Yao , H., Sun , X., Zhou , S., \& Zhang , S. (2019)
\newblock Pix2vox: Context-aware 3d reconstruction from single and multi-view images. In
\newblock \emph{2019 IEEE/CVF International Conference on Computer Vision (ICCV)}
\newblock pages 2690--2698. IEEE.

\bibitem[Yang et~al.(2025{\natexlab{a}})Yang, Sax, Liang, Henaff, Tang, Cao, Chai, Meier, and Feiszli]{yang2025fast3r}
Yang , J., Sax , A., Liang , K.~J., Henaff , M., Tang , H., Cao , A., Chai , J., Meier , F., \& Feiszli , M. (2025.
\newblock {\natexlab{a}}) Fast3r: Towards 3d reconstruction of 1000+ images in one forward pass. In
\newblock \emph{Proceedings of the Computer Vision and Pattern Recognition Conference}
\newblock pages 21924--21935.

\bibitem[Yang et~al.(2023)Yang, Zhu, Lin, Nong, and Liang]{yang2023long}
Yang , L., Zhu , Z., Lin , X., Nong , J., \& Liang , Y. (2023)
\newblock Long-range grouping transformer for multi-view 3d reconstruction. In
\newblock \emph{Proceedings of the IEEE/CVF International Conference on Computer Vision}
\newblock pages 18257--18267.

\bibitem[Yang et~al.(2025{\natexlab{b}})Yang, Liu, Zhu, Liu, Ma, Nong, and Liang]{yang2025not}
Yang , L., Liu , C., Zhu , Z., Liu , A., Ma~, H., Nong , J., \& Liang , Y. (2025.
\newblock {\natexlab{b}}) Not all frame features are equal: Video-to-4d generation via decoupling dynamic-static features. In
\newblock \emph{Proceedings of the IEEE/CVF International Conference on Computer Vision}
\newblock pages 7494--7504.

\bibitem[Yang et~al.(2026)Yang, Liu, Hu, Guan, Huang, Yi, Zhang, and Liang]{yang2026spatial}
Yang , L., Liu , J., Hu~, J., Guan , C., Huang , H., Yi~, F., Zhang , C., \& Liang , Y. (2026)
\newblock Spatial-temporal state propagation autoregressive model for 4d object generation.
\newblock \emph{arXiv preprint arXiv:2602.18830}

\bibitem[Yao et~al.(2025)Yao, Xie, Voleti, Jiang, and Jampani]{yao2025sv4d}
Yao , C.-H., Xie , Y., Voleti , V., Jiang , H., \& Jampani , V. (2025)
\newblock Sv4d 2.0: Enhancing spatio-temporal consistency in multi-view video diffusion for high-quality 4d generation. In
\newblock \emph{Proceedings of the IEEE/CVF International Conference on Computer Vision}
\newblock pages 13248--13258.

\bibitem[Yu et~al.(2025)Yu, Li, Wong, Vong, and Liang]{yu2025facnet}
Yu~, X., Li~, J., Wong , C.-C., Vong , C.-M., \& Liang , Y. (2025)
\newblock Facnet: Feature alignment fast point cloud completion network.
\newblock \emph{Computational Visual Media} {\bfseries 11}\penalty0 (1):\penalty0 141--157.

\bibitem[Yu et~al.(2026{\natexlab{a}})Yu, Liu, Qiang, Ma, Yang, Wang, Rao, and Liang]{yu2026pointcsp}
Yu~, X., Liu , A., Qiang , S., Ma~, H., Yang , L., Wang , Y., Rao , Z., \& Liang , Y. (2026.
\newblock {\natexlab{a}}) Pointcsp: Cross-sample semantic propagation and stability preservation in self-supervised point cloud learning. In
\newblock \emph{Proceedings of the IEEE/CVF Conference on Computer Vision and Pattern Recognition}
\newblock pages 10016--10026.

\bibitem[Yu et~al.(2026{\natexlab{b}})Yu, Liu, Qiang, Wang, Ma, and Liang]{yu2026pointmc}
Yu~, X., Liu , A., Qiang , S., Wang , Y., Ma~, H., \& Liang , Y. (2026.
\newblock {\natexlab{b}}) Pointmc: Multi-view consistent encoding and center-global feature fusion for point clouds understanding. In
\newblock \emph{Proceedings of the AAAI Conference on Artificial Intelligence}
\newblock \emph{40}, pp. \penalty0 12169--12177.

\bibitem[Yu et~al.(2026{\natexlab{c}})Yu, Yang, Mo, Ma, Kai, Liu, and Liang]{yu2026pointchr}
Yu~, X., Yang , L., Mo~, H., Ma~, H., Kai , F., Liu , A., \& Liang , Y. (2026.
\newblock {\natexlab{c}}) Pointchr: Point cloud analysis via curvature-aware hyperbolic rectification. In
\newblock \emph{Forty-third International Conference on Machine Learning}

\bibitem[Yuan et~al.(2025)Yuan, Wickremasinghe, Nadir, Wang, Chi, and Chan]{yuan2025seeu}
Yuan , Y., Wickremasinghe , T., Nadir , Z., Wang , X., Chi , Y., \& Chan , S.~H. (2025)
\newblock Seeu: Seeing the unseen world via 4d dynamics-aware generation.
\newblock \emph{arXiv preprint arXiv:2512.03350}

\bibitem[Zeng et~al.(2024)Zeng, Jiang, Zhu, Lu, Lin, Zhu, Hu, Cao, and Yao]{zeng2024stag4d}
Zeng , Y., Jiang , Y., Zhu , S., Lu~, Y., Lin , Y., Zhu , H., Hu~, W., Cao , X., \& Yao , Y. (2024)
\newblock Stag4d: Spatial-temporal anchored generative 4d gaussians. In
\newblock \emph{European Conference on Computer Vision}
\newblock pages 163--179. Springer.

\bibitem[Zhang et~al.(2025{\natexlab{a}})Zhang, Moing, Koppula, Rocco, Momeni, Xie, Sun, Sukthankar, Barral, Hadsell, et~al.]{zhang2025efficiently}
Zhang , C., Moing , G.~L., Koppula , S., Rocco , I., Momeni , L., Xie , J., Sun , S., Sukthankar , R., Barral , J.~K., Hadsell , R., \& others  (2025.
\newblock {\natexlab{a}})
\newblock Efficiently reconstructing dynamic scenes one d4rt at a time.
\newblock \emph{arXiv preprint arXiv:2512.08924}

\bibitem[Zhang et~al.(2024{\natexlab{a}})Zhang, Herrmann, Hur, Jampani, Darrell, Cole, Sun, and Yang]{zhang2024monst3r}
Zhang , J., Herrmann , C., Hur , J., Jampani , V., Darrell , T., Cole , F., Sun , D., \& Yang , M.-H. (2024.
\newblock {\natexlab{a}})
\newblock Monst3r: A simple approach for estimating geometry in the presence of motion.
\newblock \emph{arXiv preprint arXiv:2410.03825}

\bibitem[Zhang et~al.(2024{\natexlab{b}})Zhang, Bi, Tan, Xiangli, Zhao, Sunkavalli, and Xu]{zhang2024gs}
Zhang , K., Bi~, S., Tan , H., Xiangli , Y., Zhao , N., Sunkavalli , K., \& Xu~, Z. (2024.
\newblock {\natexlab{b}}) Gs-lrm: Large reconstruction model for 3d gaussian splatting. In
\newblock \emph{European Conference on Computer Vision}
\newblock pages 1--19. Springer.

\bibitem[Zhang et~al.(2025{\natexlab{b}})Zhang, Ge, Tian, Xu, Chen, Lv, and Shen]{zhang2025pomato}
Zhang , S., Ge~, Y., Tian , J., Xu~, G., Chen , H., Lv~, C., \& Shen , C. (2025.
\newblock {\natexlab{b}}) Pomato: Marrying pointmap matching with temporal motions for dynamic 3d reconstruction. In
\newblock \emph{Proceedings of the IEEE/CVF International Conference on Computer Vision}
\newblock pages 5680--5689.

\bibitem[Zhao et~al.(2023)Zhao, Yan, Xie, Hong, Li, and Lee]{zhao2023animate124}
Zhao , Y., Yan , Z., Xie , E., Hong , L., Li~, Z., \& Lee , G.~H. (2023)
\newblock Animate124: Animating one image to 4d dynamic scene.
\newblock \emph{arXiv preprint arXiv:2311.14603}

\bibitem[Zhao et~al.(2025)Zhao, Lai, Lin, Zhao, Liu, Yang, Feng, Yang, Zhang, Yang, et~al.]{zhao2025hunyuan3d}
Zhao , Z., Lai , Z., Lin , Q., Zhao , Y., Liu , H., Yang , S., Feng , Y., Yang , M., Zhang , S., Yang , X., \& others  (2025)
\newblock Hunyuan3d 2.0: Scaling diffusion models for high resolution textured 3d assets generation.
\newblock \emph{arXiv preprint arXiv:2501.12202}

\bibitem[Zheng et~al.(2023)Zheng, Harley, Shen, Wetzstein, and Guibas]{zheng2023pointodyssey}
Zheng , Y., Harley , A.~W., Shen , B., Wetzstein , G., \& Guibas , L.~J. (2023)
\newblock Pointodyssey: A large-scale synthetic dataset for long-term point tracking. In
\newblock \emph{Proceedings of the IEEE/CVF International Conference on Computer Vision}
\newblock pages 19855--19865.

\bibitem[Zheng et~al.(2024)Zheng, Li, Nagano, Liu, Hilliges, and De~Mello]{zheng2024unified}
Zheng , Y., Li~, X., Nagano , K., Liu , S., Hilliges , O., \& De~Mello , S. (2024)
\newblock A unified approach for text-and image-guided 4d scene generation. In
\newblock \emph{Proceedings of the IEEE/CVF Conference on Computer Vision and Pattern Recognition}
\newblock pages 7300--7309.

\bibitem[Zhu et~al.(2023{\natexlab{a}})Zhu, Yang, Li, Jiang, and Liang]{zhu2023umiformer}
Zhu , Z., Yang , L., Li~, N., Jiang , C., \& Liang , Y. (2023.
\newblock {\natexlab{a}}) Umiformer: Mining the correlations between similar tokens for multi-view 3d reconstruction. In
\newblock \emph{2023 IEEE/CVF International Conference on Computer Vision (ICCV)}
\newblock pages 18180--18189. IEEE.

\bibitem[Zhu et~al.(2023{\natexlab{b}})Zhu, Yang, Lin, Yang, and Liang]{zhu2023garnet}
Zhu , Z., Yang , L., Lin , X., Yang , L., \& Liang , Y. (2023.
\newblock {\natexlab{b}})
\newblock Garnet: Global-aware multi-view 3d reconstruction network and the cost-performance tradeoff.
\newblock \emph{Pattern Recognition} {\bfseries 142}:\penalty0 109674.

\end{thebibliography}
\normalsize


\end{document}